\documentclass[sigconf,nonacm,screen]{acmart}
\usepackage{amsmath}
\usepackage{booktabs}
\usepackage{makecell}
\usepackage{multirow}
\usepackage{tabularx}
\usepackage{array}
\usepackage{xcolor}
\usepackage{enumitem}
\usepackage{xspace}
\usepackage{placeins}

\DeclareUnicodeCharacter{2009}{\,}

\newcommand{\modelname}{UniEQA\xspace}

\title{UniEQA: Benchmarking Hierarchical EEG Understanding through a Unified Question-Answering Formulation}

\newcommand{\UniEQAZhejiang}{Zhejiang University}
\newcommand{\UniEQARinggee}{Ringgee Smart Technologies Co., Ltd.}
\newcommand{\UniEQAHKUST}{The Hong Kong University of Science and Technology (Guangzhou)}
\newcommand{\UniEQAZhejiangAffiliation}{\affiliation{\institution{\UniEQAZhejiang}\city{Hangzhou}\country{China}}}
\newcommand{\UniEQARinggeeAffiliation}{\affiliation{\institution{\UniEQARinggee}\country{China}}}
\newcommand{\UniEQAHKUSTAffiliation}{\affiliation{\institution{\UniEQAHKUST}\city{Guangzhou}\country{China}}}

\author{Yaojun Hu}
\UniEQAZhejiangAffiliation
\author{Chunlai Dong}
\UniEQAZhejiangAffiliation
\author{Chen Zhang}
\UniEQAZhejiangAffiliation
\author{Xuetian Gao}
\UniEQARinggeeAffiliation
\author{Ruijian Xu}
\UniEQARinggeeAffiliation
\author{Yuxiang Cai}
\UniEQAZhejiangAffiliation
\author{Haochao Ying}
\UniEQAZhejiangAffiliation
\author{Jian Wu}
\UniEQAZhejiangAffiliation
\author{Jintai Chen}
\UniEQAHKUSTAffiliation

\makeatletter
\if@ACM@anonymous\else
  \g@addto@macro\@authornotes{\footnotetext[1]{Co-corresponding authors.}}
  \renewcommand{\@mkauthors@iii}{%
    \global\setbox\mktitle@bx=\vbox{%
      \unvbox\mktitle@bx
      \begingroup
        \centering
        {\Large
          Yaojun Hu\textsuperscript{1}\quad
          Chunlai Dong\textsuperscript{1}\quad
          Chen Zhang\textsuperscript{1}\quad
          Xuetian Gao\textsuperscript{2}\quad
          Ruijian Xu\textsuperscript{2}\par
          Yuxiang Cai\textsuperscript{1,*}\quad
          Haochao Ying\textsuperscript{1}\quad
          Jian Wu\textsuperscript{1}\quad
          Jintai Chen\textsuperscript{3,*}\par}
        \smallskip
        {\normalsize
          \textsuperscript{1}\,\UniEQAZhejiang\qquad
          \textsuperscript{2}\,\UniEQARinggee\par
          \textsuperscript{3}\,\UniEQAHKUST\par}
      \endgroup
      \medskip
    }%
  }
\fi
\makeatother

\begin{document}

\begin{abstract}
Understanding users' intentions and states is important for personalized and interactive Web systems. Electroencephalography (EEG) captures neural activity related to intention, cognition, and affect, offering physiological evidence that can complement language and behavioral signals.
Recent EEG foundation models learn transferable representations, but their downstream application commonly uses a separate prediction head for each dataset and retains dataset-specific target spaces and protocols. Natural language question offers a shared output format, yet existing evaluations either recast dataset-specific labels through fixed prompts or confine richer question sets to short clinical EEG segments. Fine-grained comparison across EEG domains and levels of interpretation therefore remains limited. We introduce UniEQA, a question-conditioned benchmark that casts diverse EEG targets within a common question-answering framework.
UniEQA integrates 14 public datasets into \textit{1.52M} traceable question--answer pairs over \textit{575K} EEG segments. It organizes six hierarchical EEG understanding task families into three signal-to-answer abstraction levels and supports both closed-ended and open-ended evaluation on a subject-disjoint benchmark split. Experiments with signal-native EEG-language models and waveform-based vision-language models reveal distinct capability profiles across tasks and answer formats, limited zero-shot performance, and substantial gains from supervised adaptation. We will release the dataset, the construction and evaluation protocols, and the baseline results after publication.

\end{abstract}



\maketitle

\section{Introduction}

Electroencephalography (EEG), which non-invasively records brain activity with high temporal resolution, has been explored for user understanding in search and recommendation. Previous studies of reading comprehension~\citep{ye2022reading}, preference estimation~\citep{davis2021preferences}, and information-need awareness~\citep{michalkova2022awareness} highlight EEG's potential as a complementary source of information for understanding users' states and responses.
These applications differ in the user attributes and states they seek, creating heterogeneous prediction targets for EEG models. This diversity motivates a shared task formulation that preserves their distinct semantic requirements.

A natural-language interface offers a common way to query different aspects of a recording, including physiological states, task-associated attributes, and affective conditions. Evaluating this interface requires understanding which EEG targets a model can answer correctly and whether its performance benefits from the supplied signal.
However, EEG interpretation remains challenging because useful evidence is noisy, subject-dependent, distributed over channels and time, and only indirectly related to the underlying cognitive or clinical state. Application diversity adds another source of complexity, as each domain defines different targets, annotations, and levels of interpretation for the same physiological modality. General EEG understanding therefore requires models and evaluation protocols that span heterogeneous recordings and task semantics rather than a single dataset-specific label space.

\begin{figure}[t]
    \centering
    \includegraphics[width=\linewidth]{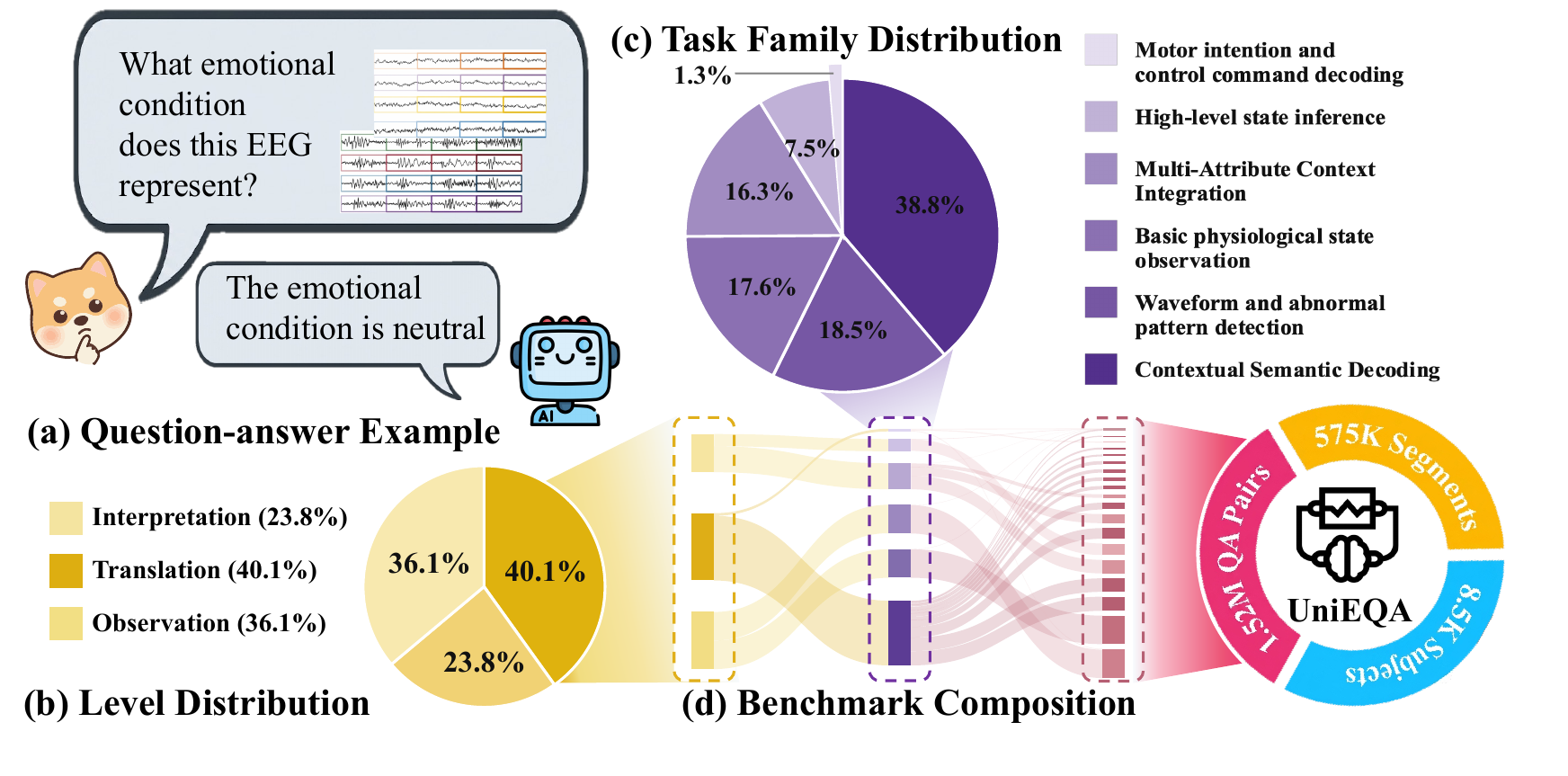}
    \caption{Overview of UniEQA. (a) Representative EEG-QA example. (b, c) Distributions across three abstraction levels and six task families. (d) Mapping from levels to task families and annotation fields, together with the benchmark scale.}
    \label{fig:intro}
\end{figure}

Recent EEG foundation models learn transferable representations from large-scale neural recordings, but their downstream performance is commonly measured through separate dataset-specific prediction tasks~\citep{jiang2024labram,yang2023biot,wang2024eegpt}. NeuralBench improves comparability across datasets and models through a standardized benchmarking framework, while retaining individual decoding tasks as its evaluation units~\citep{banville2026neuralbench}. Signal-native EEG-language models and waveform-based vision-language models further support multi-task instruction following and generative interaction, as demonstrated by NeuroLM, WaveMind, and CerebraGloss~\citep{jiang2025neurolm,zeng2025wavemind,gu2026cerebragloss}. These advances establish the feasibility of querying EEG through language. Their evaluations nevertheless address particular task collections or application domains, leaving a need for complementary assessment that organizes heterogeneous EEG targets, examines the facts recovered in generated answers, and measures performance under question-only and altered-EEG conditions.

To measure progress toward general-purpose EEG understanding, a benchmark must support both multi-task learning and fine-grained diagnosis of model capabilities.
First, multi-task learning requires heterogeneous EEG targets to share a common task format, allowing the same model to learn across task families without a separately designed predictor for each task. Natural-language questions provide this format by specifying the target semantics and answer format~\citep{hendrycks2021measuring,NEURIPS2023_6dcf277e}. Second, fine-grained diagnosis requires evaluation to distinguish the qualitatively different capabilities probed by EEG tasks. These capabilities range from identifying waveform or physiological properties, through mapping EEG activity to task-relevant semantics, to interpreting affective or clinical states and integrating multiple annotated attributes. Treating them as a flat collection obscures differences in model performance across signal observation, semantic translation, and state interpretation or attribute integration. Organizing tasks by signal-to-answer abstraction makes these target-specific capabilities separately measurable.

To meet these requirements, we introduce \textbf{\modelname}, a unified benchmark for evaluating heterogeneous EEG understanding through question answering (Figure \ref{fig:intro}). \modelname integrates 14 public EEG datasets spanning clinical, brain-computer interface, cognitive, affective, and task-state paradigms into 1,520,024 traceable question--answer pairs. The construction pipeline aligns recordings with source annotations, applies expert-defined task mappings, and validates generated questions and answers while preserving the original target semantics. Six task families are organized into three levels of target abstraction: \textit{Observation} covers waveform and physiological-state properties; \textit{Translation} covers task-associated actions, stimuli, conditions, and responses; and \textit{Interpretation} covers affective or clinical state annotations and multi-attribute compositions. UniEQA-T supports joint supervised adaptation, while UniEQA-Bench provides subject-disjoint evaluation with both closed-ended and open-ended questions.
Evaluations of signal-native EEG-language models and waveform-based vision--language models demonstrate the benchmark's value for joint multi-task learning and fine-grained capability assessment. Complementary metrics and EEG-input controls reveal factual errors, textual predictability, and task-dependent benefits from paired EEG that aggregate performance alone cannot capture. Together, these findings establish \modelname as a diagnostic benchmark for measuring what models understand about EEG and how they use it, supporting the development of EEG-based user understanding for state-aware interaction.

Our main contributions are summarized as follows:
\begin{itemize}
    \item We formulate heterogeneous EEG understanding tasks within a unified question-answering framework and organize them into a three-level signal-to-answer hierarchy comprising six task families, supporting joint training with a single model without task-specific prediction heads.
    \item We construct \modelname, a large-scale EEG-QA benchmark that transforms 14 heterogeneous public EEG datasets across clinical, BCI, cognitive, affective, and task-state domains into 1,520,024 traceable question--answer pairs over 575,845 EEG segments.
    \item We systematically evaluate representative signal-native EEG-language and waveform-based vision--language models across abstraction levels, task families, and answer formats, revealing capability profiles and localizing model limitations in observation, translation, and interpretation.
\end{itemize}

\section{Related Work}
\label{sec:related}

\noindent\textbf{EEG for user understanding.}
EEG provides complementary evidence about users' cognitive and affective responses that may not be fully reflected in explicit ratings. Research in this area serves two related purposes. For characterizing information processing, studies examine neural responses associated with reading comprehension and information-need awareness~\cite{ye2022reading,michalkova2022awareness}. For supporting interactive services, EEG-derived relevance and preference estimates inform document re-ranking and collaborative filtering~\cite{ye2024brainfeedback,davis2021preferences}. This application-oriented line also connects EEG with multidimensional engagement in short-video recommendation~\cite{zhang2024eegsvrec} and uses multimodal physiological information to generate natural-language intervention text~\cite{zhang2025interveeg}. Together, these studies illustrate how EEG can inform both the interpretation of user states and the design of system responses. Their evaluations are tailored to application-specific outcomes, motivating complementary assessment across heterogeneous EEG targets. \modelname supports this assessment through a shared question-answering interface that evaluates answer correctness and dependence on correctly paired signals.

\noindent\textbf{EEG foundation models.}
Early methods introduced contrastive and masked-prediction objectives for transferable EEG representation learning~\cite{kostas2021bendr,chien2022maeeg,foumani2024eeg2rep}. Subsequent foundation models extend this direction through heterogeneous biosignal tokenization, large-scale pretraining, and spatiotemporal or spatial modeling~\cite{yang2023biot,jiang2024labram,wang2024eegpt,wang2025cbramod,doner2025luna,elouahidi2025reve,fang2025neuript,xiao2025brainomni}. These models improve transfer across datasets, but transfer is generally evaluated by adapting a separate predictor to each dataset's native label space, segmentation protocol, and metric. NeuralBench substantially advances comparability by standardizing data access and evaluation across 14 architectures, 36 tasks, and 94 datasets~\cite{banville2026neuralbench}. Its basic evaluation unit, however, remains an individual downstream decoding task with a task-specific target space. \modelname complements this effort by expressing heterogeneous EEG targets through a shared question-answering format and organizing them within a common signal-to-answer hierarchy.

\noindent\textbf{EEG-language modeling.}
EEG-language research has progressed from cross-modal alignment and text decoding to instruction-based interaction. EEG-CLIP and EEG-Language Pretraining align clinical recordings with reports to support label-efficient classification and EEG-text retrieval~\cite{camaretndir2025eegclip,gijsen2025eeglanguage}. DeWave translates reading-evoked EEG into text, while Thought2Text generates descriptions of visual stimuli from EEG~\cite{duan2023dewave,mishra2025thought2text}. More recent systems extend language interaction across multiple EEG tasks. NeuroLM represents EEG as discrete neural tokens and unifies several downstream datasets through instruction tuning~\cite{jiang2025neurolm}. WaveMind supports descriptions, multiple-choice questions, and open-ended interaction across four downstream tasks~\cite{zeng2025wavemind}, while CerebraGloss develops waveform-based description, QA, and dialogue for clinical EEG interpretation~\cite{gu2026cerebragloss}. These studies establish the feasibility of language-conditioned EEG question-answer modeling, but their evaluation suites remain coupled to particular training corpora, task collections, or application domains. Rather than introducing another EEG-language architecture, \modelname focuses on the evaluation: it provides a model-independent protocol that organizes heterogeneous targets into three abstraction levels and six task families, enabling direct comparison between signal-native and waveform-based models.

\section{UniEQA}
\label{sec:unieqa_dataset}

\begin{figure*}[t]
    \centering
    \includegraphics[width=\textwidth]{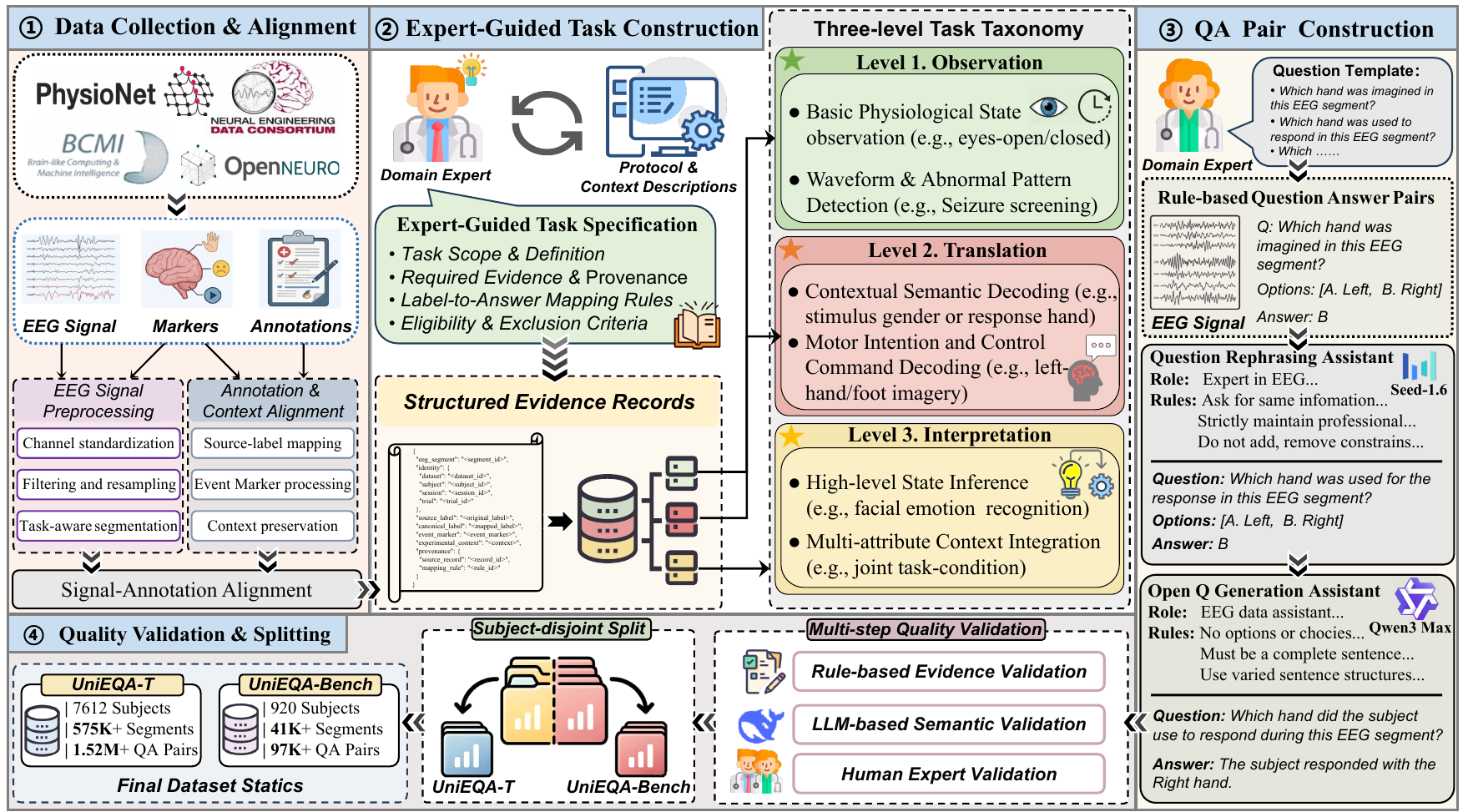}
    \caption{Overview of UniEQA construction. EEG signals and annotations are transformed into structured evidence records, organized by an expert-defined task hierarchy, converted into QA pairs, and validated before subject-disjoint partitioning.}
    \label{fig:workflow}

\end{figure*}

This section presents the hierarchical task taxonomy and the four-stage pipeline used to construct UniEQA, as summarized in Figure~\ref{fig:workflow}. The pipeline first aligns heterogeneous EEG signals with their source annotations, then applies expert-defined task specifications to produce structured evidence records. The records are then converted into rule-based, paraphrased, and open-ended question--answer pairs, followed by multi-stage quality validation and subject-disjoint splitting into UniEQA-T and UniEQA-Bench.

\subsection{Hierarchical Task Taxonomy}
To structure heterogeneous EEG targets, we organize six task families into three levels of target abstraction: \textit{Observation}, \textit{Translation}, and \textit{Interpretation}. The hierarchy progresses from waveform and physiological-state descriptions, through task-associated semantics, to affective or clinical state annotations and multi-attribute compositions. Task assignments follow the meaning and composition of the source annotations and their context.

\noindent\textbf{Observation-level tasks} identify information that is directly grounded in EEG with target answers remaining close to the physiological recordings. The relevant evidence comes from waveform morphology, rhythmic structure, spectral signatures, and stable background organization. Consequently, these tasks require little semantic abstraction and rely primarily on signal-level patterns.
This level comprises two tasks: \textit{Waveform and Abnormal Pattern Detection}, which focuses on clinically salient local events (e.g., epileptiform discharges, slowing, and abnormal transients)~\cite{kuhlmann2018seizure}, and \textit{Basic Physiological State Observation}, which addresses broader signal-grounded conditions such as sleep stage, vigilance, and baseline physiological state~\cite{adamantidis2019oscillating}. The former emphasizes detection of localized waveform events, while the latter emphasizes recognition of broader physiological organization over time.

\noindent\textbf{Translation-level tasks} aim to map EEG dynamics to semantic targets that are not directly readable from the signal. Unlike observation-level tasks, the central challenge at this level is semantic mapping rather than direct signal inspection. Therefore, the answer cannot be determined from waveform structure alone. Instead, it must be inferred by linking distributed neural activity to a task label, behavioral intention, or contextual meaning. This level comprises two tasks: \textit{Motor Intention and Control Command Decoding}, common in brain--computer interface settings, maps sensorimotor activity to internally generated action semantics such as intended movement or control commands~\cite{chaudhary2016bci}. In contrast, \textit{Contextual Semantic Decoding} links EEG responses to externally defined semantics such as stimulus category and task condition. The key distinction between the two lies in the origin of the target semantics: endogenous intention in the former versus exogenous experimental context in the latter.

\noindent\textbf{Interpretation-level tasks} address affective or clinical states and composite descriptions associated with an EEG recording. Their placement at this level reflects the broader semantic scope or multi-attribute composition of the target. This level comprises two tasks: \textit{High-Level State Inference}, which predicts source-provided affective or clinical state annotations, including emotion categories~\cite{alarcao2019emotions} and diagnostic groups, and \textit{Multi-attribute Context Integration}, which combines multiple annotated attributes associated with the same signal window into one answer. The former targets a state-level interpretation, whereas the latter targets a composite description assembled from source-linked attributes.

The hierarchy organizes prediction targets by abstraction and composition without prescribing the internal reasoning process used by a model. More abstract targets may present greater empirical challenges. Option count, answer length, and the number of required factual slots characterize answer-space and response complexity; empirical difficulty also depends on source imbalance, signal quality, model interface, and target identifiability within the selected window. We therefore examine performance across both abstraction levels and individual task families.

\subsection{Dataset Construction Pipeline}
\label{sec:benchmark}

To reformulate heterogeneous EEG supervision under a common question-conditioned interface, we develop a construction pipeline that combines fidelity, linguistic diversity, and structured task coverage for fine-grained understanding capability evaluation.
As illustrated in Figure~\ref{fig:workflow}, the pipeline aligns recordings with annotations, applies expert-defined mappings to the task taxonomy, and validates the resulting question--answer pairs before subject-disjoint partitioning into UniEQA-T and UniEQA-Bench.
Table~\ref{tab:data_distribution} summarizes the distributions of segments, subjects, and QA pairs across the six tasks in both splits.

\noindent\textbf{Data Collection and Alignment. }
To cover the six tasks, we curate 14 public EEG datasets spanning clinical, BCI, cognitive, affective, and task-state domains, including BCICIV~\cite{blankertz2007non,tangermann2012review}, GAL~\cite{luciw2014multi}, CHB-MIT~\cite{guttag2010chb}, DEMENTIA~\cite{ds004504:1.0.8}, SEED~\cite{zheng2015seed}, and TUSZ~\cite{shah2018temple,obeid2016temple}. The collection covers representative applications such as motor imagery, emotion recognition, and seizure detection.
Operationally, a source can enter the construction pipeline only when its recordings can be associated with source-qualified subject identifiers, its relevant temporal units can be aligned with ground-truth annotations, and its label definitions admit a deterministic mapping to one or more task targets. These requirements preserve traceability from every benchmark answer to a native annotation while allowing sources with different acquisition protocols to share one question-conditioned interface.
For each source, we process EEG signals and their annotations through two coordinated streams.
Source-specific signal converters extract task-relevant temporal units and retain the identifiers needed to locate their parent recordings. Construction records preserve ordered channel names and source sampling-rate metadata when these fields are available. Preprocessing follows each source protocol rather than imposing one filter chain on all recordings. All model-ready signal files used in the reported experiments are stored at 200~Hz.
In parallel, the annotation stream establishes source-label mappings, processes event markers to align each segment with its original annotation, and preserves the corresponding acquisition and task context. Subject and session identifiers and source provenance are retained throughout. Together, these operations produce harmonized signal-annotation records for expert-guided task construction.

\begin{table}[t]
\centering
\setlength{\tabcolsep}{4pt}
\caption{Validation results on the same instances from each task family across LLM and human experts. LLM, E1, and E2 report the pass rates of DeepSeek-V4-Pro and two independent EEG experts. L-E1, L-E2, and E1-E2 report exact agreement between the corresponding binary validity judgments.}
\begin{tabular}{@{}lcccccc@{}}
\toprule
\textbf{Task}
& \textbf{LLM}
& \textbf{E1}
& \textbf{E2}
& \textbf{L-E1}
& \textbf{L-E2}
& \textbf{E1-E2} \\
\midrule
Abnormal Pattern   & 98.7 & 98.4 & 98.6 & 99.1 & 99.3 & 99.4 \\
Physiol. State     & 98.1 & 97.8 & 98.0 & 98.9 & 99.1 & 99.4 \\
Motor Intention    & 97.4 & 97.1 & 97.3 & 98.7 & 98.9 & 99.2 \\
Context. Semantics & 97.0 & 96.7 & 96.9 & 98.5 & 98.7 & 99.2 \\
High-Level State   & 95.1 & 94.8 & 95.0 & 98.1 & 98.3 & 98.8 \\
Multi-attribute    & 94.3 & 94.0 & 94.2 & 97.7 & 97.9 & 98.6 \\
\midrule
\textbf{Overall}   & \textbf{96.8} & \textbf{96.5}
                   & \textbf{96.7} & \textbf{98.5}
                   & \textbf{98.7} & \textbf{99.1} \\
\bottomrule
\end{tabular}
\label{tab:llm_human_validation}
\end{table}

\begin{table*}[t]
\centering
\setlength{\tabcolsep}{4pt}
\caption{Distributions of EEG segments, subjects, and question-answer (QA) pairs in UniEQA-T and UniEQA-Bench.
}
\begin{tabular}{cc|rrrrrr|r}
\toprule
\multirow{2}{*}{\textbf{Split}} & \multicolumn{1}{c|}{\textbf{Level}}
& \multicolumn{2}{c}{\textbf{Observation-level}}
& \multicolumn{2}{c}{\textbf{Translation-level}}
& \multicolumn{2}{c|}{\textbf{Interpretation-level}}
& \multirow{2}{*}{\textbf{Overall}} \\
\cmidrule(lr){3-4}\cmidrule(lr){5-6}\cmidrule(lr){7-8}
& \multicolumn{1}{c|}{\textbf{Task}}
& \shortstack{\textbf{Abnormal}\\\textbf{Pattern}}
& \shortstack{\textbf{Physio.}\\\textbf{State}}
& \shortstack{\textbf{Contextual}\\\textbf{Semantic}}
& \shortstack{\textbf{Motor}\\\textbf{Intention}}
& \shortstack{\textbf{High-level}\\\textbf{State}}
& \shortstack{\textbf{Multi-attri.}\\\textbf{Context}}
\\
\midrule

\multirow{3}{*}{\parbox{2.6cm}{\centering\textbf{UniEQA-Bench}}}
& \multicolumn{1}{c|}{EEG Segments}
& 12,584 & 6,049  & 7,217 & 1,405 & 8,290 & 5,886 & 41,431 \\
& \multicolumn{1}{c|}{Subjects}
& 870 & 15 & 10 & 2 & 15 & 8 & 920 \\
& \multicolumn{1}{c|}{QA Pairs}
& 25,168 & 12,140 & 29,062 & 2,810 & 16,580 & 11,772 & 97,532 \\
\midrule

\multirow{3}{*}{\parbox{2.6cm}{\centering\textbf{UniEQA-T}}}
& \multicolumn{1}{c|}{EEG Segments}
& 120,896 & 117,581 & 123,325 & 8,423 & 47,720 & 116,469 & 534,414 \\
& \multicolumn{1}{c|}{Subjects}
& 6,753 & 289 & 208 & 12 & 188 & 162 & 7,612 \\
& \multicolumn{1}{c|}{QA Pairs}
& 241,792 & 254,114 & 576,538 & 21,646 & 95,437 & 232,965 & 1,422,492 \\
\bottomrule
\end{tabular}

\label{tab:data_distribution}

\end{table*}

\noindent\textbf{Expert-Guided Task Construction. }
Based on the aligned signals and ground-truth annotations, EEG domain experts review the task protocol and available ground-truth annotation fields, then assign each field to one of the six task families and its corresponding abstraction level based on target provenance and composition. Waveform and physiological-state annotations map to observation, annotations of a single trial-level action, stimulus, task condition, or behavioral response map to translation, and higher-level clinical or affective annotations and multi-attribute compositions map to interpretation. The experts further define deterministic label-to-answer mappings and valid answer vocabularies. These task specifications are combined with the aligned signal, source annotation, acquisition context, and provenance metadata to form a Structured Evidence Record. All questions generated from this record inherit its task and level assignments.

For example, GAL contributes recording-state annotations to Basic Physiological State Observation, object weight and surface friction to Contextual Semantic Decoding, and state--weight--friction compositions to Multi-attribute Context Integration. These targets use distinct native annotation fields while retaining their linkage to the same source recording.

\noindent{\textbf{Structured Evidence Records.}}
The {Structured Evidence Record} is the logical source-linked representation formed by joining a compact QA row with its associated signal record and native annotations. The compact row stores the source task, source-qualified subject identifier, signal-file key, abstraction level, target attribute, question type, question, options, answer, and task family. Associated signal records retain source-specific fields such as channel names, sampling rate, trial or event identifiers, and segment boundaries when available. Expert task specification adds the selected annotation fields, canonical target, label-to-answer mapping, and valid answer vocabulary. This separation keeps language fields distinct from the evidence used to derive their targets without requiring every source to expose an identical physical metadata schema.

\noindent\textbf{Question-answer Pair Construction.}
Starting from the Structured Evidence Records, we construct QA pairs that preserve the same source-grounded targets while varying their linguistic realization and answer format.

\noindent{\emph{Rule-Based Question-Answer Seed Construction.}}
For the five single-field task families, a domain-expert-authored template verbalizes the selected annotation field and a fixed mapping produces the ground-truth answer. Dataset-specific converters read the native annotation fields from each pre-segmented record and discard missing, empty, \texttt{N/A}, or out-of-vocabulary values before question construction. Each retained value is mapped to a fixed answer vocabulary for the corresponding source task. Multiple-choice distractors are drawn only from that vocabulary. The converters deterministically permute the options using a global seed of 42 together with the signal-file key and target attribute, then recompute the correct option letter after permutation.

\noindent{\emph{Question rephrasing.}}
Fixed templates may allow models to exploit recurring syntactic patterns. We therefore use Seed-1.6~\cite{team2025seed1} to paraphrase a selected subset of questions with high template redundancy. The model rewrites only the question text, while the canonical answer, answer options, task assignment, and source linkage remain unchanged. The prompt requires each paraphrase to query exactly the same target without adding, removing or altering factual constraints.

\noindent{\emph{Multi-attribute Question Integration.}}
Multi-attribute Context Integration composes two or more targets only when all constituent annotations are valid and refer to the same signal window. For Multi-attribute Context Integration, Qwen3-Max~\citep{yang2025qwen3} receives the same-window subquestions, their option sets, and the resolved answer text for each constituent attribute. This stage changes the linguistic presentation of the composed target but does not introduce a target that is absent from the constituent annotations.

\noindent{\emph{Open-ended generation.}}
We further use Qwen3-Max~\cite{yang2025qwen3} to derive open-ended question-answer pairs from the closed-ended instances. Each generation prompt contains the original question, options, and canonical answer. The generated question must remain grounded in the same source annotation, and its answer must preserve the original target without introducing unsupported clinical or cognitive information. The generation prompt also encourages linguistic variation beyond the original template. Here, open-ended refers to the response format: models generate an answer without selecting from a supplied option list. This setting evaluates the recovery and verbalization of source-linked facts, including multiple attributes where applicable, rather than unrestricted EEG-language reasoning.

\noindent\textbf{Quality Validation. } To ensure the question-answer pairs are faithful to the original evidence, we perform multi-stage quality control before inclusion in the final benchmark.

\noindent{\emph{Rule-based evidence validation.}}
To ensure that generated answers preserve all source-derived facts, we perform deterministic fact-slot validation for every instance. For each source task and annotation field, we define the required factual slots, admissible values, and lexical rules covering label names, synonymous expressions, and negation patterns. After normalization, these rules extract sets of \textit{(slot, value)} facts from the generated and ground-truth answers. Shared facts are counted as true positives, unsupported facts as false positives, and omitted reference facts as false negatives. An instance passes only when all required slots are recovered without conflicting or unsupported values. For multi-attribute answers, this criterion must hold jointly across all constituent slots. We additionally verify segment-annotation alignment, option validity, output format, and required provenance fields.

\noindent{\emph{LLM-based semantic validation.}}
To identify semantic errors not captured by deterministic fact-slot checks, we use DeepSeek-V4-Pro~\cite{xu2026deepseek} as a rubric-based evaluator. Given the Structured Evidence Record together with the original and generated question--answer pairs, the evaluator assesses evidence consistency, semantic preservation, and the absence of answer leakage. Each dimension receives a binary score of 0 or 1 under a fixed scoring prompt. An instance passes semantic validation only when all three dimensions receive a score of 1, corresponding to a total score of 3. All remaining instances are excluded. The complete scoring rubric and prompt are provided in the supplementary material.

\noindent{\emph{Human expert validation.}}
To assess residual errors and the reliability of the automated auditor, two EEG-domain experts independently apply the same rubric and decision rule to a task-stratified sample of 6,000 instances, with 1,000 instances randomly sampled from each task family. For each instance, the experts assign a binary pass/fail judgment based on four criteria: consistency with the source annotation, answer correctness, preservation of the intended task semantics, and the absence of unsupported clinical or cognitive interpretations. Table~\ref{tab:llm_human_validation} shows comparable overall pass rates across the LLM evaluator and two experts, with all pairwise agreements exceeding 98\%. This correspondence with expert judgments supports the reliability of the LLM-based semantic validation.

\noindent\textbf{Subject-disjoint splitting.}
We partition the validated records at the subject level within each source dataset using source-specific subject allocation rules. SEED assigns its first 12 subjects to UniEQA-T and the remaining three to UniEQA-Bench, while TUSZ retains the provider-defined training and development partitions for UniEQA-T and the evaluation partition for UniEQA-Bench. The remaining sources use fixed source-local subject assignments, documented in the supplementary material. All recordings, aligned windows, and derived QA variants associated with the same source-qualified subject remain in one split. An audit of the frozen exports confirms zero source-qualified subject overlap between UniEQA-T and UniEQA-Bench.

\noindent\textbf{Data Statistics. }
Final UniEQA comprises two complementary subsets: UniEQA-T for large-scale training, and UniEQA-Bench for held-out evaluation. UniEQA-T contains 1,422,492 question-answer pairs over 534,414 EEG segments, whereas UniEQA-Bench contains 97,532 pairs over 41,431 segments. As detailed in Table~\ref{tab:data_distribution}, each subset covers all three abstraction levels and six task families, supporting fine-grained capability evaluation beyond a single aggregate score.

\section{Experiments}

\begin{table*}[t]
\centering
\setlength{\tabcolsep}{0.5pt}
\caption{Task-wise closed- and open-ended results on UniEQA-Bench. Closed-ended cells report option-selection accuracy and label macro-F1 as Acc. and M-F1. Open-ended questions report ROUGE-1 (R-1), BERTScore (BS), and fact-slot F1 (S-F1).  Z.s. and F.t. denote zero-shot evaluation and evaluation after fine-tuning, respectively. }
\label{tab:results-combined}
\begin{tabular}{@{}
l@{\hspace{3pt}}c@{\hspace{3pt}}
c@{\hspace{3.2pt}}c@{\hspace{3.2pt}}c@{\hspace{10pt}}
c@{\hspace{3.2pt}}c@{\hspace{3.2pt}}c@{\hspace{10pt}}
c@{\hspace{3.2pt}}c@{\hspace{3.2pt}}c@{\hspace{10pt}}
c@{\hspace{3.2pt}}c@{\hspace{3.2pt}}c@{\hspace{10pt}}
c@{\hspace{3.2pt}}c@{\hspace{3.2pt}}c@{\hspace{10pt}}
c@{\hspace{3.2pt}}c@{\hspace{3.2pt}}c
@{}}
\toprule
\textbf{Model} & \textbf{Setting}
& \multicolumn{3}{c@{\hspace{10pt}}}{\makebox[0pt][c]{\textbf{Abnorm. Pat.}}}
& \multicolumn{3}{c@{\hspace{10pt}}}{\makebox[0pt][c]{\textbf{Physiol. State}}}
& \multicolumn{3}{c@{\hspace{10pt}}}{\makebox[0pt][c]{\textbf{Motor Intent.}}}
& \multicolumn{3}{c@{\hspace{10pt}}}{\makebox[0pt][c]{\textbf{Context. Sem.}}}
& \multicolumn{3}{c@{\hspace{10pt}}}{\makebox[0pt][c]{\textbf{High-L. State}}}
& \multicolumn{3}{c}{\makebox[0pt][c]{\textbf{Multi-Attr. Ctx.}}} \\

\midrule
\multicolumn{20}{c}{\itshape Closed-ended evaluation} \\
\midrule
& & \multicolumn{3}{c@{\hspace{10pt}}}{Acc.\quad M-F1}
& \multicolumn{3}{c@{\hspace{10pt}}}{Acc.\quad M-F1}
& \multicolumn{3}{c@{\hspace{10pt}}}{Acc.\quad M-F1}
& \multicolumn{3}{c@{\hspace{10pt}}}{Acc.\quad M-F1}
& \multicolumn{3}{c@{\hspace{10pt}}}{Acc.\quad M-F1}
& \multicolumn{3}{c}{Acc. \quad M-F1} \\
\midrule
NeuroLM-B & F.t.
& \multicolumn{3}{c@{\hspace{10pt}}}{97.3 \quad 60.5}
& \multicolumn{3}{c@{\hspace{10pt}}}{78.3 \quad 60.7}
& \multicolumn{3}{c@{\hspace{10pt}}}{62.3 \quad 48.3}
& \multicolumn{3}{c@{\hspace{10pt}}}{58.2 \quad 34.5}
& \multicolumn{3}{c@{\hspace{10pt}}}{54.2 \quad 37.6}
& \multicolumn{3}{c}{71.9 \quad 62.0} \\
NeuroLM-L & F.t.
& \multicolumn{3}{c@{\hspace{10pt}}}{97.2 \quad 49.1}
& \multicolumn{3}{c@{\hspace{10pt}}}{79.1 \quad 65.5}
& \multicolumn{3}{c@{\hspace{10pt}}}{70.2 \quad 60.0}
& \multicolumn{3}{c@{\hspace{10pt}}}{53.1 \quad 31.2}
& \multicolumn{3}{c@{\hspace{10pt}}}{40.0 \quad 32.3}
& \multicolumn{3}{c}{64.5 \quad 59.6} \\
NeuroLM-XL & F.t.
& \multicolumn{3}{c@{\hspace{10pt}}}{{97.4 \quad 61.2}}
& \multicolumn{3}{c@{\hspace{10pt}}}{{80.0 \quad 66.3}}
& \multicolumn{3}{c@{\hspace{10pt}}}{{71.0 \quad 61.0}}
& \multicolumn{3}{c@{\hspace{10pt}}}{{59.0 \quad 35.2}}
& \multicolumn{3}{c@{\hspace{10pt}}}{{55.0 \quad 38.3}}
& \multicolumn{3}{c}{{72.6 \quad 62.8}} \\
\midrule
Qwen3-VL-2B & Z.s.
& \multicolumn{3}{c@{\hspace{10pt}}}{68.9 \quad 25.1}
& \multicolumn{3}{c@{\hspace{10pt}}}{2.9 \quad 29.4}
& \multicolumn{3}{c@{\hspace{10pt}}}{42.7 \quad 35.0}
& \multicolumn{3}{c@{\hspace{10pt}}}{40.1 \quad 24.3}
& \multicolumn{3}{c@{\hspace{10pt}}}{38.7 \quad 26.0}
& \multicolumn{3}{c}{34.9 \quad 36.8} \\
Qwen3-VL-2B & F.t.
& \multicolumn{3}{c@{\hspace{10pt}}}{97.3 \quad 48.8}
& \multicolumn{3}{c@{\hspace{10pt}}}{88.5 \quad 70.4}
& \multicolumn{3}{c@{\hspace{10pt}}}{75.2 \quad 60.3}
& \multicolumn{3}{c@{\hspace{10pt}}}{67.4 \quad 40.7}
& \multicolumn{3}{c@{\hspace{10pt}}}{62.3 \quad 40.9}
& \multicolumn{3}{c}{85.2 \quad 74.6} \\
InternVL2-2B & Z.s.
& \multicolumn{3}{c@{\hspace{10pt}}}{36.6 \quad 18.3}
& \multicolumn{3}{c@{\hspace{10pt}}}{3.4 \quad 28.1}
& \multicolumn{3}{c@{\hspace{10pt}}}{47.8 \quad 38.8}
& \multicolumn{3}{c@{\hspace{10pt}}}{41.0 \quad 24.6}
& \multicolumn{3}{c@{\hspace{10pt}}}{35.6 \quad 28.1}
& \multicolumn{3}{c}{33.9 \quad 31.6} \\
InternVL2-2B & F.t.
& \multicolumn{3}{c@{\hspace{10pt}}}{97.3 \quad 52.2}
& \multicolumn{3}{c@{\hspace{10pt}}}{87.9 \quad 76.9}
& \multicolumn{3}{c@{\hspace{10pt}}}{76.3 \quad 63.1}
& \multicolumn{3}{c@{\hspace{10pt}}}{66.6 \quad 39.1}
& \multicolumn{3}{c@{\hspace{10pt}}}{59.3 \quad 40.7}
& \multicolumn{3}{c}{85.3 \quad 63.9} \\
GLM-Edge-V2 & Z.s.
& \multicolumn{3}{c@{\hspace{10pt}}}{50.2 \quad 21.1}
& \multicolumn{3}{c@{\hspace{10pt}}}{37.1 \quad 24.5}
& \multicolumn{3}{c@{\hspace{10pt}}}{43.4 \quad 38.3}
& \multicolumn{3}{c@{\hspace{10pt}}}{42.2 \quad 27.2}
& \multicolumn{3}{c@{\hspace{10pt}}}{38.0 \quad 25.6}
& \multicolumn{3}{c}{36.1 \quad 41.7} \\
GLM-Edge-V2 & F.t.
& \multicolumn{3}{c@{\hspace{10pt}}}{97.2 \quad 53.1}
& \multicolumn{3}{c@{\hspace{10pt}}}{87.2 \quad 73.5}
& \multicolumn{3}{c@{\hspace{10pt}}}{73.2 \quad 58.7}
& \multicolumn{3}{c@{\hspace{10pt}}}{67.4 \quad 41.4}
& \multicolumn{3}{c@{\hspace{10pt}}}{63.0 \quad 42.8}
& \multicolumn{3}{c}{84.3 \quad 70.2} \\
CerebraGloss-3B & Z.s.
& \multicolumn{3}{c@{\hspace{10pt}}}{61.2 \quad {23.7}}
& \multicolumn{3}{c@{\hspace{10pt}}}{8.1 \quad {28.6}}
& \multicolumn{3}{c@{\hspace{10pt}}}{28.8 \quad {30.2}}
& \multicolumn{3}{c@{\hspace{10pt}}}{29.4 \quad {21.7}}
& \multicolumn{3}{c@{\hspace{10pt}}}{37.0 \quad {25.8}}
& \multicolumn{3}{c}{32.0 \quad {33.4}} \\
CerebraGloss-3B & F.t.
& \multicolumn{3}{c@{\hspace{10pt}}}{97.2 \quad 51.8}
& \multicolumn{3}{c@{\hspace{10pt}}}{84.1 \quad 64.8}
& \multicolumn{3}{c@{\hspace{10pt}}}{73.1 \quad 59.8}
& \multicolumn{3}{c@{\hspace{10pt}}}{65.1 \quad 36.3}
& \multicolumn{3}{c@{\hspace{10pt}}}{61.5 \quad 42.7}
& \multicolumn{3}{c}{84.1 \quad 70.8} \\
Qwen2.5-VL-7B & Z.s.
& \multicolumn{3}{c@{\hspace{10pt}}}{{62.0 \quad 24.2}}
& \multicolumn{3}{c@{\hspace{10pt}}}{{9.0 \quad 29.0}}
& \multicolumn{3}{c@{\hspace{10pt}}}{{29.5 \quad 30.8}}
& \multicolumn{3}{c@{\hspace{10pt}}}{{30.2 \quad 22.3}}
& \multicolumn{3}{c@{\hspace{10pt}}}{{37.8 \quad 26.5}}
& \multicolumn{3}{c}{{33.0 \quad 34.0}} \\
Qwen2.5-VL-7B & F.t.
& \multicolumn{3}{c@{\hspace{10pt}}}{{97.4 \quad 53.0}}
& \multicolumn{3}{c@{\hspace{10pt}}}{{85.0 \quad 66.0}}
& \multicolumn{3}{c@{\hspace{10pt}}}{{74.0 \quad 61.0}}
& \multicolumn{3}{c@{\hspace{10pt}}}{{66.0 \quad 38.0}}
& \multicolumn{3}{c@{\hspace{10pt}}}{{62.0 \quad 43.5}}
& \multicolumn{3}{c}{{85.0 \quad 71.5}} \\
\midrule
\multicolumn{20}{c}{\itshape Open-ended evaluation} \\
\midrule
& & R-1 & BS & S-F1
& R-1 & BS & S-F1
& R-1 & BS & S-F1
& R-1 & BS & S-F1
& R-1 & BS & S-F1
& R-1 & BS & S-F1 \\
\midrule
NeuroLM-B & F.t.
& 87.9 & 97.8 & 95.3 & 91.9 & 98.6 & 76.0 & 82.9 & 96.6 & 63.3
& 85.5 & 97.7 & 58.9 & 30.6 & 89.8 & 47.7 & 84.9 & 97.8 & 60.5 \\
NeuroLM-L & F.t.
& 87.2 & 97.7 & 94.8 & 91.0 & 98.5 & 75.6 & 82.6 & 96.5 & 65.6
& 85.2 & 97.4 & 54.7 & 24.4 & 88.2 & 33.6 & 83.5 & 97.5 & 55.9 \\
NeuroLM-XL & F.t.
& {88.3} & {97.9} & {95.6}
& {92.1} & {98.7} & {76.8}
& {83.4} & {96.8} & {66.2}
& {85.8} & {97.8} & {59.6}
& {31.5} & {90.0} & {48.5}
& {85.2} & {97.9} & {61.3} \\
\midrule
Qwen3-VL-2B & Z.s.
& 14.1 & 85.9 & 22.4 & 8.8 & 83.6 & 4.7 & 1.4 & 82.0 & 0.0
& 7.8 & 83.6 & 12.7 & 5.4 & 84.9 & 3.8 & 12.2 & 83.5 & 2.4 \\
Qwen3-VL-2B & F.t.
& 87.5 & {97.8} & 94.9 & 91.8 & {98.4} & 85.8 & 85.9 & {97.0} & 78.6
& 87.6 & {97.8} & 67.8 & 30.4 & {89.5} & 52.6 & 86.6 & {98.0} & 72.6 \\
InternVL2-2B & Z.s.
& 41.1 & 90.0 & 20.0 & 33.4 & 88.3 & 19.4 & 61.6 & 92.1 & 0.0
& 57.8 & 91.7 & 25.8 & 13.0 & 85.1 & 12.8 & 23.6 & 86.4 & 7.3 \\
InternVL2-2B & F.t.
& 87.5 & 97.8 & 95.2 & 91.3 & 98.6 & 83.6 & 85.5 & 97.1 & 77.9
& 87.6 & 98.0 & 67.0 & 30.8 & 89.7 & 53.2 & 86.7 & 98.1 & 72.6 \\
GLM-Edge-V2 & Z.s.
& 18.3 & 86.2 & 14.2 & 21.9 & 86.0 & 16.6 & 26.7 & 87.6 & 0.0
& 42.1 & 89.6 & 25.9 & 6.8 & 83.1 & 4.6 & 20.5 & 85.7 & 3.8 \\
GLM-Edge-V2 & F.t.
& 87.2 & 97.7 & 94.5
& 91.2 & 98.5 & 84.4
& 85.2 & 97.0 & 73.3
& 87.6 & 98.0 & 68.4
& 27.9 & 88.3 & 55.5
& 86.5 & 98.0 & 72.1 \\
CerebraGloss-3B & Z.s.
& 18.5 & 87.2 & 3.2 & 2.1 & 72.3 & {0.6} & 5.2 & 75.8 & {0.8}
& 8.3 & 78.1 & {2.7} & 11.7 & 81.2 & {3.4} & 12.3 & 79.4 & {3.1} \\
CerebraGloss-3B & F.t.
& 86.8 & 97.7 & 94.8 & 90.5 & 98.5 & 80.8 & 83.9 & 96.8 & 74.8
& 87.6 & 97.9 & 67.0 & 31.3 & 89.8 & 55.0 & 85.7 & 98.0 & 69.2 \\
Qwen2.5-VL-7B & Z.s.
& {20.0} & {88.0} & {4.0}
& {3.0} & {74.0} & {1.0}
& {6.0} & {77.0} & {1.2}
& {9.0} & {79.0} & {3.2}
& {12.5} & {82.0} & {4.0}
& {13.0} & {80.5} & {3.8} \\
Qwen2.5-VL-7B & F.t.
& {87.2} & {97.8} & {95.0}
& {90.8} & {98.5} & {81.5}
& {84.2} & {96.9} & {75.5}
& {87.8} & {97.9} & {67.5}
& {31.8} & {89.9} & {55.8}
& {86.0} & {98.0} & {70.0} \\
\bottomrule
\end{tabular}
\end{table*}

\subsection{Experimental Setup}
We evaluate question-conditioned EEG understanding under a shared task and evaluation protocol, selecting models that support natural-language questions and answers for both closed- and open-ended tasks.
Signal-native models include NeuroLM-B, NeuroLM-L, and NeuroLM-XL~\citep{jiang2025neurolm}, which directly encode multichannel EEG using their native signal-processing and optimization protocols. Waveform-based models follow the CerebraGloss paradigm~\citep{gu2026cerebragloss}, rendering each segment as a channel-stacked waveform image. This family includes CerebraGloss-3B, Qwen3-VL-2B~\citep{bai2025qwen3vl}, InternVL2-2B~\citep{chen2024far}, Qwen2.5-VL-7B~\citep{bai2025qwen25vltechnicalreport}, and GLM-Edge-V2~\citep{zhipu2024glmedge}.

\noindent\textbf{Evaluation Metrics.}
Closed-ended predictions are parsed into the benchmark answer vocabulary and evaluated by exact option-selection accuracy and canonical semantic-label macro-F1. Invalid or unparseable selections count as incorrect. Accuracy measures overall correctness. Semantic-label macro-F1 gives equal weight to the frozen canonical labels and is therefore more informative under class imbalance. It is distinct from macro-F1 over option positions, whose meanings can change when the same semantic label appears at different positions.
Open-ended predictions are evaluated by ROUGE-1~\citep{lin2004rouge}, BERTScore F1~\citep{zhang2020bertscore}, and task-specific fact-slot F1. ROUGE-1 measures unigram overlap and BERTScore measures contextual semantic similarity. Fact-slot evaluation applies the frozen task schema to predictions and references, accumulates true positives, false positives, and false negatives over extracted facts, and reports the resulting F1. Because a generic EEG response can obtain semantic similarity without recovering the required source-linked facts, lexical, semantic, and slot metrics are interpreted jointly.

Detailed model configurations and additional results are provided in the supplementary material.

\subsection{Unified EEG Understanding  Evaluation}

Table~\ref{tab:results-combined} reports zero-shot and fine-tuned results for closed- and open-ended QA across six task families, enabling analysis of joint adaptation, task-level diagnosticity, model-family profiles, and metric complementarity.

\noindent\textbf{Joint Adaptation across Tasks.}
Across the five waveform-based models with matched zero-shot and fine-tuned results, fine-tuning improves accuracy, ROUGE-1, and fact-slot F1 on every task. The gains span target types and answer formats, demonstrating that UniEQA-T provides effective joint supervision through a shared question-conditioned formulation. These consistent gains show that UniEQA-T supports joint adaptation of a single model across the full task space.

\noindent\textbf{Task-wise Diagnostic Resolution.}
Fine-tuned models generally achieve higher closed-ended accuracy on observation-level tasks than on translation-level tasks and High-Level State Inference. This pattern suggests that moving from signal-level descriptions toward task semantics and higher-level state targets often presents greater  challenges under the evaluated protocols. However, the relationship is not monotonic: Multi-attribute Context Integration achieves comparatively strong performance despite belonging to the interpretation level, and it exceeds High-Level State Inference in open-ended fact-slot F1 for every fine-tuned model. These results suggest that target abstraction is informative for characterizing task difficulty, while source composition, label distributions, and target identifiability also affect performance. The hierarchy therefore captures a broad progression in target abstraction, with task-wise evaluation revealing both common performance trends and exceptions.

\noindent\textbf{EEG Input Representation and Task Performance. } Signal-native and waveform-based models show task-dependent differences. For  abnormal pattern detection task, signal-native models achieve the highest macro-F1 and a slight advantage in open-ended fact-slot F1. Waveform-based models lead on most metrics across the other five tasks, particularly contextual semantic decoding and multi-attribute context integration. These patterns suggest an interaction between EEG representation and pretrained model capabilities, although architectural and training differences prevent isolating the effect of input format.

\noindent\textbf{Complementarity of Answer Formats and Metrics.} Metric choice materially affects the interpretation of model performance. On abnormal pattern detection, fine-tuned accuracy exceeds 97\%, while macro-F1 remains below 62\%, showing that accuracy can mask limited class-balanced performance under label imbalance. In open-ended evaluation, high BERTScore can coexist with substantially lower fact-slot F1, indicating that semantic similarity does not ensure factual recovery. The metrics are therefore complementary, distinguishing overall correctness, class-balanced recognition, semantic similarity, and recovery of the required facts.

\begin{table}[t]
\centering
\setlength{\tabcolsep}{2pt}
\caption{Question-only controls relative to full-input performance on UniEQA-Bench. Question-only inference omits EEG only at evaluation, whereas question-only SFT omits it during both training and evaluation.}
\label{tab:eeg_grounding}
\begin{tabular*}{\linewidth}{@{\extracolsep{\fill}}llccc@{}}
\toprule
\multirow{2}{*}{\textbf{Model}} &
\multirow{2}{*}{\textbf{Metric}} &
\multicolumn{2}{c}{\textbf{Question-only}} &
\multirow{2}{*}{\makecell{\textbf{Full}\\\textbf{Input}}} \\
\cmidrule(lr){3-4}
& & \textbf{Inference} & \textbf{SFT} & \\
\midrule
\multirow{4}{*}{NeuroLM-B}
& Accuracy               & 62.53 & 66.58 & 71.88 \\
& Macro-F1             & 37.20 & 41.75 & 48.30 \\
& BERTScore F1           & 95.54 & 96.05 & 96.47 \\
& Slot F1                & 54.30 & 59.16 & 67.99 \\
\addlinespace[0.5pt]
\multirow{4}{*}{NeuroLM-L}
& Accuracy               & 63.31 & {64.92} & 67.37 \\
& Macro-F1             & 39.25 & {40.86} & 46.42 \\
& BERTScore F1           & 95.59 & {95.74} & 96.03 \\
& Slot F1                & 57.63 & {58.91} & 63.59 \\
\midrule
\multirow{4}{*}{Qwen3-VL-2B}
& Accuracy               & 38.99 & {64.83} & 79.24 \\
& Macro-F1             & 30.58 & {41.18} & 54.98 \\
& BERTScore F1           & 83.00 & {95.92} & 96.62 \\
& Slot F1                &  6.10 & {59.74} & 75.26 \\
\addlinespace[0.5pt]
\multirow{4}{*}{InternVL2-2B}
& Accuracy               & 35.27 & {63.58} & 78.48 \\
& Macro-F1             & 28.96 & {40.37} & 53.39 \\
& BERTScore F1           & 86.11 & {95.88} & 96.57 \\
& Slot F1                &  9.58 & {58.96} & 74.81 \\
\addlinespace[0.5pt]
\multirow{4}{*}{CerebraGloss-3B }
& Accuracy               & 34.19 & 64.15 & 77.68 \\
& Macro-F1             & 30.92 & 39.50 & 51.98 \\
& BERTScore F1           & 84.52 & 95.99 & 96.52 \\
& Slot F1                &  5.42 & 58.23 & 73.81 \\
\bottomrule
\end{tabular*}

\end{table}
\subsection{EEG Dependence Beyond Textual Priors}

\noindent\textbf{Question-Only Controls.}
Question-only inference removes the EEG or waveform input only at evaluation while preserving the trained multimodal model, question, options, target, split, and answer normalization. It is therefore a pathway-sensitivity control under an input-distribution shift. Question-only SFT removes the EEG or image from every training and evaluation batch and passes only input tokens, attention masks, and answer labels. This matched-modality control measures predictability from question wording, answer spaces, and supervision without EEG.

\noindent\textbf{Textual Priors Analysis.}
Table~\ref{tab:eeg_grounding} compares Full Input with two complementary question-only controls. Removing EEG only at inference reduces accuracy by between 40.25 and 43.49 percentage points for Qwen3-VL-2B, InternVL2-2B, and CerebraGloss-3B, compared with between 4.06 and 9.35 points for NeuroLM-B and NeuroLM-L. These reductions reflect sensitivity to modality removal under an input-distribution shift and therefore do not isolate the contribution of instance-specific EEG evidence. Question-only SFT provides a matched-modality comparison and achieves substantial performance, indicating that question wording, answer spaces, and training supervision support prediction without EEG. Full Input nevertheless improves accuracy by between 13.53 and 14.90 percentage points and Slot F1 by between 15.52 and 15.85 points across the three waveform-based models. For NeuroLM-B, the corresponding gains are smaller, at 5.30 and 8.83 points, respectively. These results support an additional predictive benefit from EEG under the evaluated protocols while exposing substantial textual predictability within the current benchmark within the current benchmark within the current benchmark. Full-input scores should therefore be interpreted together with question-only and mismatched-EEG controls, rather than treated alone as a measure of instance-specific EEG understanding.

\noindent\textbf{Prediction Shift under EEG Substitution.}
To verify whether model predictions are conditioned on the supplied EEG, we hold the question and answer options constant and replace each paired recording with a label-disjoint EEG drawn from the same source task. From the full test set, we retain closed-ended instances with recoverable, label-disjoint original and substituted labels under the same target schema and a unique option match for the substituted label. As shown in Table~\ref{tab:donor_following}, substituted-target accuracy exceeds original-target accuracy for all six models and macro-F1 follows the same ordering, although the two scores are nearly identical for NeuroLM-L. Because the textual input, model, and inference procedure remain unchanged, this systematic alignment with the substituted label indicates that predictions respond to information associated with the supplied EEG rather than remaining anchored to the original answer or a question--answer prior.

\begin{table}[t]
\centering
\setlength{\tabcolsep}{3pt}
\caption{Comparison of model-prediction agreement with substituted-EEG and original-target reference labels under label-disjoint replacement.}
\label{tab:donor_following}
\begin{tabular}{@{}lcccc@{}}
\toprule
\textbf{Model} & \multicolumn{2}{c}{\textbf{Substituted Target}} & \multicolumn{2}{c}{\textbf{Original Target}} \\
\cmidrule(lr){2-3}\cmidrule(lr){4-5}
& \textbf{Accuracy} & \textbf{Macro-F1} & \textbf{Accuracy} & \textbf{Macro-F1} \\
\midrule
NeuroLM-B       & 60.40 & 31.36 & 28.13 & 28.03 \\
NeuroLM-L       & 50.12 & 31.46 & 34.82 & 31.34 \\
\midrule
CerebraGloss-3B & 69.37 & 37.86 & 23.06 & 26.20 \\
Qwen3-VL-2B     & 69.95 & 39.73 & 22.40 & 29.43 \\
InternVL2-2B    & 69.39 & 36.47 & 22.82 & 27.63 \\
GLM-Edge-V2     & 70.04 & 37.92 & 22.83 & 26.80 \\
\bottomrule
\end{tabular}
\end{table}

\begin{figure}[!t]
    \centering
    \includegraphics[width=0.95\columnwidth]{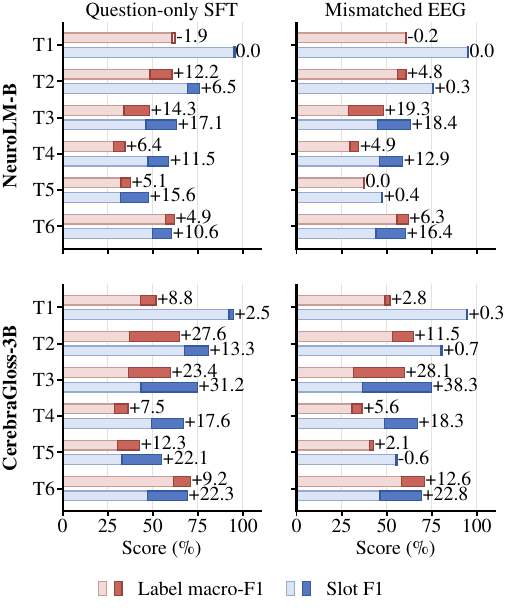}
    \caption{EEG dependence varies across tasks and model interfaces. Dark extensions denote gains from full input. T1--T6 follow the task order in Table~\ref{tab:results-combined}.}\label{fig:eeg_grounding_task_gains}
    \vspace{-4mm}
\end{figure}

\noindent\textbf{Task-Specific Dependence.}
Mismatched EEG keeps all target text fixed and deterministically rotates complete signal groups within the same task. The mapping is a derangement with no same-signal or cross-task pairing, but labels are not required to differ.
Figure~\ref{fig:eeg_grounding_task_gains} compares Full Input with Question-only SFT and Mismatched EEG to assess task-specific sensitivity to modality removal and incorrect EEG pairing.
Motor Intention and Control Command Decoding shows that the largest gains over Mismatched EEG for both models and both metrics, together with substantial improvements over Question-only SFT. Paired EEG also yields substantial Slot F1 gains for Contextual Semantic Decoding and Multi-attribute Context Integration, which recover or compose record-linked semantic attributes. In contrast, Waveform and Abnormal Pattern Detection and High-Level State Inference show limited sensitivity to mismatching, indicating that paired-input dependence does not increase mechanically with abstraction level. CerebraGloss-3B is generally more sensitive than NeuroLM-B to both modality removal and incorrect pairing, particularly for physiological state, motor intention, and multi-attribute integration. These task-specific and model-dependent patterns show that \modelname distinguishes where and to what extent model performance benefits from correctly paired EEG under the evaluated protocols.

\FloatBarrier

\section{Conclusion}

We introduced UniEQA, a benchmark for hierarchical EEG understanding through a unified question-answering formulation. UniEQA contains 1.52M traceable question--answer pairs over 575K EEG segments and organizes six hierarchical task families across observation, translation, and interpretation, with closed-ended, open-ended, and subject-disjoint evaluation. Experiments with signal-native EEG-language and waveform-based vision-language models show limited zero-shot performance and consistent gains from supervised adaptation across all task.
The results further reveal task-dependent differences across levels, EEG input representations, and answer formats. Complementary metrics expose class-balanced and factual errors hidden by accuracy or semantic similarity, while question-only and substituted-EEG controls show that sensitivity to correctly paired EEG varies across tasks and models.
By covering motor intentions, task contexts, and affective states alongside broader physiological and clinical targets, UniEQA provides a common testbed for this assessment and a resource for investigating the potential of EEG-assisted Web interaction.

\bibliographystyle{ACM-Reference-Format}
\bibliography{references}
\par
\nobalance
\begingroup
\appendix
\setcounter{figure}{0}
\setcounter{table}{0}
\renewcommand{\thefigure}{S\arabic{figure}}
\renewcommand{\thetable}{S\arabic{table}}
\renewcommand{\theHfigure}{supp.\arabic{figure}}
\renewcommand{\theHtable}{supp.\arabic{table}}
\setcounter{secnumdepth}{2}
\setlength{\emergencystretch}{1.5em}
\setlength{\textfloatsep}{8pt plus 2pt minus 2pt}
\setlength{\dbltextfloatsep}{8pt plus 2pt minus 2pt}
\setlength{\floatsep}{7pt plus 2pt minus 2pt}
\setlength{\dblfloatsep}{7pt plus 2pt minus 2pt}
\setlength{\intextsep}{7pt plus 2pt minus 2pt}
\renewcommand{\topfraction}{0.95}
\renewcommand{\dbltopfraction}{0.95}
\renewcommand{\bottomfraction}{0.85}
\renewcommand{\textfraction}{0.05}
\setcounter{topnumber}{3}
\setcounter{totalnumber}{5}
\captionsetup{skip=4pt}
\raggedbottom
\definecolor{promptblue}{RGB}{36,104,165}
\definecolor{promptgreen}{RGB}{38,132,91}
\definecolor{promptorange}{RGB}{191,102,35}
\newcolumntype{Y}{>{\raggedright\arraybackslash}X}
\newsavebox{\unieqapromptcontent}
\newenvironment{promptbox}[2][]{%
  \def\unieqapromptcolor{#2}%
  \setlength{\fboxsep}{4pt}%
  \setlength{\fboxrule}{0.6pt}%
  \begin{lrbox}{\unieqapromptcontent}%
  \begin{minipage}{\dimexpr\linewidth-2\fboxsep-2\fboxrule\relax}%
  \colorbox{#2!80!black}{%
    \parbox{\dimexpr\linewidth-2\fboxsep\relax}{%
      \color{white}\small\bfseries #1}}\par\smallskip
}{%
  \end{minipage}%
  \end{lrbox}%
  \noindent\fcolorbox{\unieqapromptcolor!80!black}{\unieqapromptcolor!5}{%
    \usebox{\unieqapromptcontent}}\par
}

\section{Prompt Templates and Construction Interfaces}
\label{app:qa_prompts}

Figures~\ref{fig:question_rephrasing_template}, \ref{fig:open_ended_generation_template}, and~\ref{fig:qc_prompt_template} summarize the input fields, output schemas, and semantic constraints used by the language-generation and validation stages. These condensed specifications document the intended transformations and acceptance criteria rather than reproducing verbatim service requests. The generation specifications vary surface form or answer format while keeping the source-derived target fixed, and the validation rubric assigns separate binary judgments to evidence consistency, semantic preservation, and answer leakage.

\begin{table*}[t]
\centering
\small
\setlength{\tabcolsep}{3.5pt}
\renewcommand{\arraystretch}{0.96}
\begin{tabularx}{\textwidth}{@{}>{\raggedright\arraybackslash}p{0.095\textwidth}Y>{\raggedright\arraybackslash}p{0.20\textwidth}>{\raggedright\arraybackslash}p{0.18\textwidth}@{}}
\toprule
\textbf{Source} & \textbf{Full name or description} & \textbf{Hosting platform} & \textbf{Access terms} \\
\midrule
\textbf{BCICIV} & BCI Competition IV Dataset 1 calibration set & BBCI competition site & Competition terms \\
\textbf{GAL} & WAY-EEG-GAL: multi-channel EEG recordings during grasp-and-lift trials & Figshare collection & CC0 1.0 \\
\textbf{CHB-MIT} & CHB-MIT Scalp EEG Database & PhysioNet, version 1.0.0 & ODC-By 1.0 \\
\textbf{DEMENTIA} & EEG recordings from Alzheimer's disease, frontotemporal dementia, and healthy groups & OpenNeuro, ds004504 & CC0 1.0 \\
\textbf{SEED} & SJTU Emotion EEG Dataset & SJTU BCMI Laboratory & Academic-only access \\
\textbf{TUSZ} & TUH EEG Seizure Corpus & Temple University and NEDC & Credentialed access \\
\midrule
\textbf{ASYM} & EEG-Asymmetries Dataset & OpenNeuro, ds007172 & CC0 1.0 \\
\textbf{EXO} & EEG-Controlled Exoskeleton for Walking and Standing & OpenNeuro, ds007788 & CC0 1.0 \\
\textbf{IS} & Synchronous EEG and fMRI dataset on inner speech & OpenNeuro, ds006033 & CC0 1.0 \\
\textbf{MOM} & Multiple Object Monitoring (EEG) & OpenNeuro, ds006768 & CC0 1.0 \\
\textbf{PHOTO} & Open-eyes photostimulation EEG from AD, FTD, and healthy groups & OpenNeuro, ds006036 & CC0 1.0 \\
\textbf{SURPRISE} & Surprise-induced failures of visual and auditory awareness & OpenNeuro, ds006394 & CC0 1.0 \\
\textbf{TDCS} & TDCS Modulation of Visual Cortex in Motor Imagery & OpenNeuro, ds006126 & CC0 1.0 \\
\textbf{VRNF} & EEG recordings during sham neurofeedback in virtual reality & OpenNeuro, ds006839 & CC0 1.0 \\
\bottomrule
\end{tabularx}
\caption{Source information and access conditions for the 14 public EEG datasets in UniEQA. Provider-specific and credentialed-access terms are not open-source licenses and remain governed by the original providers.}
\label{tab:dataset_licenses}
\end{table*}

\begin{figure}[!htbp]
\centering
\begin{promptbox}[Question Rephrasing Specification]{promptblue}
\small
\textbf{Role.} You are a senior clinical neurophysiologist and an expert in electroencephalography (EEG), with knowledge of EEG waveform morphology, frequency bands, clinical interpretation, 10--20/10--10 montages, and artifact identification. Your task is to rephrase the given question while strictly preserving its original clinical and technical meaning.

\textbf{Rules.}
\begin{enumerate}[leftmargin=*,itemsep=1pt,topsep=2pt]
    \item The rephrased question must ask for exactly the same information, so the original answer and options remain correct.
    \item Use a different sentence structure, vocabulary, or phrasing while preserving professional EEG terminology, including terms such as spikes, sharp waves, rhythmic activity, frequency bands, and montage names.
    \item Do not add, remove, or change factual constraints. For example, do not change ``this segment'' to ``this trial'' or ``epoch'' if these words are not equivalent in the source record.
    \item Output only the rephrased question text, with no quotes and no explanation.
\end{enumerate}

\textbf{Input.}
\texttt{Original question: \{question\}} \\
\texttt{Options: \{options\}}

\textbf{Output.}
\texttt{Rephrased question}
\end{promptbox}
\caption{Question-rephrasing specification. Only the wording of a closed-ended question changes; its options and answer remain fixed.}
\label{fig:question_rephrasing_template}
\Description{Question-rephrasing specification with a role, four rules, input fields, and the required output.}
\end{figure}

\begin{figure}[!htbp]
\centering
\begin{promptbox}[Open-Ended Generation Specification]{promptgreen}
\small
\textbf{Role.} Rewrite a structured EEG event description into a natural-language question--answer pair while preserving the source-derived target.

\textbf{Rules.}
\begin{enumerate}[leftmargin=*,itemsep=1pt,topsep=2pt]
    \item The question must be open-ended and descriptive. Do not ask yes/no questions and do not include options or binary choices.
    \item The answer must be a complete sentence strictly based on the provided \texttt{Answer} value.
    \item Do not introduce any new medical interpretation, diagnosis, cognitive-state inference, or task information beyond the input.
    \item Do not repeat or list the provided options in the output.
    \item Use varied sentence structures across samples.
    \item Output only valid JSON in the specified format, with no extra text.
\end{enumerate}

\textbf{Input JSON.}
\begin{quote}\small\ttfamily\raggedright
\{"Question": "...", "options": ["...", "...", "..."],\\
"Answer": "..."\}
\end{quote}

\textbf{Output JSON.}
\begin{quote}\small\ttfamily\raggedright
\{"Question": "...", "Answer": "..."\}
\end{quote}
\end{promptbox}
\caption{Input--output specification for open-ended generation. The transformation converts a structured instance into a question and complete-sentence answer while preserving the source-derived target.}
\label{fig:open_ended_generation_template}
\Description{Open-ended generation specification with six target-preservation rules and input and output JSON schemas.}
\end{figure}

\begin{figure}[t]
\centering
\begin{promptbox}[Semantic Validation Rubric]{promptorange}
\small
\sloppy
\textbf{Role.} You are an EEG-QA quality-control auditor. Given a source EEG metadata record and a generated QA pair, determine whether the QA pair is faithful to the source annotation, answerable from the intended evidence field, and assigned to the correct UniEQA level and task family.

\textbf{Binary rubric.}
\begin{enumerate}[leftmargin=*,itemsep=1pt,topsep=2pt]
    \item \textbf{Evidence consistency:} the answer must be derivable from the provided source label or structured evidence field.
    \item \textbf{Semantic preservation:} the generated wording must preserve the original clinical, cognitive, behavioral, or experimental meaning, use terminology consistently with the source record, retain the declared task assignment, and introduce no unsupported interpretation.
    \item \textbf{Absence of answer leakage:} copied target labels, option text, and dataset-specific identifiers must not reveal the answer anywhere in the wording of the question.
\end{enumerate}

Each dimension receives 1 only when its criterion is fully satisfied. The instance passes only when all three dimensions receive 1.

\textbf{Input JSON.}
\begin{quote}\small\ttfamily\raggedright
\{"source\_dataset": "...", "level": "...", "task": "...",\\
"evidence": \{...\}, "Question": "...", "options": [...],\\
"Answer": "...", "generated\_question": "...",\\
"generated\_answer": "..."\}
\end{quote}

\textbf{Output JSON.}
\begin{quote}\small\ttfamily\raggedright
\{"evidence\_consistency": 0 | 1,\\
"semantic\_preservation": 0 | 1,\\
"no\_answer\_leakage": 0 | 1,\\
"rationale": "..."\}
\end{quote}
\end{promptbox}
\caption{Rubric and output schema for LLM-based semantic validation. An instance passes only when all three binary criteria receive a score of one.}
\label{fig:qc_prompt_template}
\Description{Semantic-validation rubric covering evidence consistency, semantic preservation, and absence of answer leakage, with JSON schemas.}
\end{figure}

\subsection{LLM Usage Statement}
\label{app:llm_usage}
LLMs support data construction without creating new EEG labels. Seed-1.6 paraphrases a selected subset of questions with high template redundancy. Qwen3-Max integrates same-window attribute questions into composite multiple-choice instances and separately generates open-ended QA pairs. DeepSeek-V4-Pro applies the semantic-validation rubric. Source annotations and canonical targets remain fixed throughout: models may alter only permitted language fields or return validation decisions. Figures~\ref{fig:question_rephrasing_template}--\ref{fig:qc_prompt_template} specify semantic equivalence, the prohibition of unsupported interpretations, and output schemas.

\section{Dataset Sources and Subject Splits}
\label{app:datasets}

Table~\ref{tab:dataset_licenses} summarizes the source datasets and their access conditions.

\subsection{Subject-Disjoint Splitting}
\label{app:subject_splits}
All recordings, aligned windows, and derived QA variants associated with a source-qualified subject are assigned to the same split. SEED assigns the first 12 source subjects to UniEQA-T and the remaining three to UniEQA-Bench. TUSZ retains the provider-defined train and development partitions on the training side and the evaluation partition on the benchmark side. The remaining held-out source-local identifiers are ASYM (13, 19, 21), BCICIV (6), CHB-MIT (12, 24), DEMENTIA (34--36, 63--65, 86--88), EXO (2), GAL (11), IS (2), MOM (4, 5), PHOTO (6, 46, 61), SURPRISE (16, 18), TDCS (5), and VRNF (2, 8). Subject-level checks confirm that UniEQA-T and UniEQA-Bench share no source-qualified subjects, including no shared TUSZ patient identifiers.

\section{Additional Experimental Details}
\label{app:detailed_experiment_results}
\label{app:benchmark_eval}

\subsection{Benchmark Protocol and Model Interfaces}
\label{app:experiment_protocol}
\label{app:benchmark_construction}

\paragraph{Evaluation split.}
UniEQA-Bench is a held-out, source-qualified subject-disjoint split. All recordings and derived questions for one subject remain together, preventing cross-split overlap in that subject's spectral structure, channel characteristics, recording conditions, and repeated trials. The benchmark contains 97,532 QA pairs over 41,431 EEG segments, with 29,220 distinct signal-file keys after cross-task deduplication. Closed- and open-ended formats each contribute 48,766 pairs. UniEQA-T provides 1,422,492 pairs for supervised adaptation; zero-shot models receive none.

\paragraph{Common question interface.}
All models receive the same natural-language question and, for closed-ended items, the same ordered answer options. Signal-native NeuroLM-B, NeuroLM-L, and NeuroLM-XL encode the multichannel signal through their native EEG interface~\citep{jiang2025neurolm}. Qwen3-VL-2B~\citep{bai2025qwen3vl}, InternVL2-2B~\citep{chen2024far}, Qwen2.5-VL-7B~\citep{bai2025qwen25vltechnicalreport}, GLM-Edge-V2~\citep{zhipu2024glmedge}, and CerebraGloss-3B~\citep{gu2026cerebragloss} instead receive a channel-stacked waveform rendering of the same segment. Because the two model families also differ in architecture, pretraining, and optimization, their scores characterize benchmark behavior but do not isolate a causal effect of input representation.

\paragraph{Waveform rendering.}
For the waveform-based family, each signal undergoes per-channel DC removal followed by fixed source-level amplitude scaling. Renderings use 200-$\mu$V vertical offsets, one-second time ticks, a 200-$\mu$V calibration bar, and at most 10 seconds of displayed signal without temporal stretching or padding. Signals with at least eight recognized standard channels are mapped to a fixed 19-channel average-reference montage. Otherwise, up to 19 native channels are retained. The decoded rendering is resized without cropping or padding, preserving aspect ratio and limiting the longer edge to 384 pixels, and stored as a lossless PNG.

\balance
\subsection{Training and Evaluation Details}

\paragraph{Signal-native adaptation.}
NeuroLM-B contains 255,107,712 parameters, of which 169,597,440 are trainable, and is adapted for five epochs with global batch size 64. NeuroLM-L contains 501,331,072 total parameters with 415,820,800 trainable parameters and is adapted for one epoch. Both use AdamW with $\beta=(0.9,0.95)$, an initial learning rate of $5\times10^{-4}$, cosine decay to $5\times10^{-5}$, a 0.1 warmup ratio and weight decay 0.1.

\paragraph{Waveform-VLM adaptation.}
CerebraGloss-3B, Qwen3-VL-2B, InternVL2-2B, and GLM-Edge-V2 are adapted on the full training set for one epoch, corresponding to 22,227 optimization steps. Training uses BF16, maximum sequence length 2,048, global batch size 64, LoRA rank 8, LoRA scale 32, dropout 0.05, learning rate $10^{-4}$, cosine scheduling, a 0.05 warmup ratio, weight decay 0.1, and seed and data seed 1337. The language model is frozen and LoRA is applied to the vision encoder and multimodal aligner. No held-out validation split is used for model selection, so evaluation uses the model state after the final training step. Trainable adapter counts are 5,714,944 (CerebraGloss-3B), 3,604,480 (Qwen3-VL-2B), 3,227,648 (InternVL2-2B), and 4,664,320 (GLM-Edge-V2).

\paragraph{Inference.}
NeuroLM evaluation uses greedy decoding with a batch size of 64, at most 8 newly generated tokens for closed-ended questions, and at most 64 for open-ended questions. Waveform-VLM evaluation uses deterministic greedy decoding with temperature 0, one beam, at most 64 new tokens, maximum sequence length 2,048, and seed 1337. Variable visual-token lengths require singleton inference batches for the waveform-VLM family. Invalid or unparseable responses are retained and scored as incorrect rather than silently discarded.

\subsection{EEG Input Controls}
\label{app:question_only_controls}
\label{app:input_controls}
Question-only inference (Omit) removes EEG only at evaluation, retaining the model trained with full input; this measures pathway sensitivity under an input-distribution shift. Question-only SFT omits EEG during both training and evaluation, measuring predictability from question wording, answer spaces, and supervision alone. Blank replaces EEG with an uninformative input while preserving the interface.

Within-task shuffling deterministically rotates complete signal groups while keeping target text fixed. It permits neither same-signal nor cross-task pairings, but labels may coincide. The main paper's label-disjoint substitution instead uses the same source task, recoverable original and replacement labels under a common target schema, and a unique option match for the replacement label. Post-hoc comparison of existing shuffled predictions with replacement and original labels differs from a new substitution inference run.

\section{Additional Results}

\subsection{Complete GLM-Edge-V2 Input Ablation}

Table~\ref{tab:glm_input_ablation} reports the complete four-condition GLM-Edge-V2 input ablation. All four conditions are evaluated on the same 97,532 benchmark instances. Full input exceeds modality removal, blank input, and within-task shuffling on all reported metrics. The Full--Shuffle difference supports dependence on the correctly paired EEG representation under this protocol. It does not by itself isolate a purely electrophysiological causal mechanism, and these results do not quantify variation across independent training runs.

\begin{table}[htbp]
\centering
\small
\setlength{\tabcolsep}{3pt}
\renewcommand{\arraystretch}{1.08}
\begin{tabular}{@{}lrrrr@{}}
\toprule
\textbf{Metric} & \textbf{Full} & \textbf{Omit} & \textbf{Blank} & \textbf{Shuffle} \\
\midrule
Closed accuracy & 79.04 & 36.75 & 51.94 & 69.31 \\
Closed label macro-F1 & 55.21 & 30.44 & 32.60 & 48.11 \\
All-format label macro-F1 & 49.84 & 23.03 & 24.80 & 42.80 \\
Normalized EM & 76.15 & 20.46 & 30.69 & 66.24 \\
Slot precision & 75.62 & 31.57 & 34.61 & 64.81 \\
Slot recall & 74.85 & 8.81 & 10.06 & 64.04 \\
Slot F1 & 75.23 & 13.78 & 15.59 & 64.42 \\
\bottomrule
\end{tabular}
\caption{GLM-Edge-V2 input ablation on the complete UniEQA-Bench split. Values are percentages; EM denotes exact match after answer normalization. Shuffle uses the within-task derangement described in Section~\ref{app:input_controls}.}
\label{tab:glm_input_ablation}
\end{table}

\FloatBarrier

\endgroup

\end{document}